%% file: main.tex
\PassOptionsToPackage{table}{xcolor}
\documentclass[10pt, a4paper]{article}

\usepackage{lrec-coling2024} 

\usepackage[markup=underlined]{changes}
\usepackage{multirow}
\usepackage{import}
\usepackage{subfig}
\usepackage{boldline}

\usepackage{amssymb}
\usepackage{pifont}
\usepackage[export]{adjustbox}

\title{OATS: \textit{O}pinion \textit{A}spect \textit{T}arget \textit{S}entiment Quadruple Extraction Dataset for Aspect-Based Sentiment Analysis}

\name{Siva Uday Sampreeth Chebolu, Franck Dernoncourt, Nedim Lipka, Thamar Solorio} 

\address{University of Houston, Adobe Research, Adobe Research, University of Houston \\
         4800 Calhoun Rd, Houston, TX 77004, USA, 325 Park Ave San Jose, CA 95110, USA, \\
         325 Park Ave San Jose, CA 95110, USA, 4800 Calhoun Rd, Houston, TX 77004, USA \\
         sivauday.sampreeth8@gmail.com, franck.dernoncourt@gmail.com, \\ 
         lipka@adobe.com, thamar.solorio@gmail.com\\
         }

\abstract{
    Aspect-based sentiment analysis (ABSA) delves into understanding sentiments specific to distinct elements within 
    a user-generated review.
    It aims to analyze user-generated reviews to determine a) the target entity being reviewed, b) the high-level aspect to which it belongs, c) the sentiment words used to express the opinion, and d) the sentiment expressed toward the targets and the aspects. 
    While various benchmark datasets have fostered advancements in ABSA, they often come with domain limitations and data granularity challenges. Addressing these, we introduce the OATS dataset, which encompasses three fresh domains and consists of 27,470 sentence-level quadruples and 17,092 review-level tuples. Our initiative seeks to bridge specific observed gaps: the recurrent focus on familiar domains like restaurants and laptops, limited data for intricate quadruple extraction tasks, and an occasional oversight of the synergy between sentence and review-level sentiments. Moreover, to elucidate OATS's potential and shed light on various ABSA subtasks that OATS can solve, we conducted experiments, establishing initial baselines.
    We hope the OATS dataset augments current resources, paving the way for an encompassing exploration of ABSA (\url{https://github.com/RiTUAL-UH/OATS-ABSA}).
 \\ \newline  
}

\begin{document}

\maketitleabstract

\section{Introduction}
\label{Sec: Introduction}

\input{Sections/1-Introduction}

\section{Related Works}
\label{Sec: Related-Works}
\input{Sections/2-Related-Works}

\section{OATS Dataset}
\label{Sec: Dataset}

\input{Sections/3-Dataset}

\section{Experiments}
\label{Sec: Experiments}
\input{Sections/4-Experiments}

\section{Results and Discussion}
\label{Sec: Discussion}
\input{Sections/5-Discussion}

\section{Significance of OATS}
\label{Sec: Significance}

\input{Sections/5-Significance}


\section{Conclusion}
\label{Sec: Conclusion}
\input{Sections/6-Conclusion}

\section{Limitations and Ethical Considerations}
\label{Sec: Ethics}
\input{Sections/8-Ethics}


\section{Bibliographical References}\label{sec:reference}

\bibliographystyle{lrec-coling2024-natbib}
\bibliography{main}

\bibliographystylelanguageresource{main}

\end{document}

%% file: Sections/1-Introduction.tex
    The trend in analyzing online reviews has shifted from extracting a broad understanding of consumer opinions on overall product performance to a more granular examination of individual product aspects. This shift demands a different approach to analyzing reviews. Aspect-based sentiment Analysis (ABSA) emerged as the answer to this nuanced requirement, focusing on sentiment pertaining to specific aspects of an entity \cite{Hu-2004}. However, current datasets often fall short of capturing the complete spectrum of ABSA. The main goal of ABSA is to identify the target of the opinion, the aspect category it belongs to, the opinion phrase, and the sentiment polarity associated with the opinion. One of the significant limitations is the inability to perform joint detection of all ABSA elements due to the absence of at least one critical component in the human-annotated reviews. This limitation stunts the potential of ABSA tasks. 
    Despite the SemEval datasets' popularity, many of their sentences often group multiple aspects under a single sentiment polarity. As noted by \citeauthor{MAMS-dataset} (\citeyear{MAMS-dataset}), this approach simplifies the ABSA task, reducing it to mere sentence-level sentiment analysis. 
    As a solution, they proposed a new large-scale Multi-Aspect Multi-Sentiment dataset, where each phrase has at least two independent aspects with different sentiment polarity. However, their introduction of a "miscellaneous" category or neutral sentiment when a sentence contains only one opinion tuple is impractical. This makes the task more challenging, but it is not realistic.
    
    Recently released ABSA datasets such as ACOS \cite{ACOS} and ASQP \cite{ASQP} provide comprehensive annotations for all four elements. However, they are largely limited to the long-standing traditional domains of restaurants and laptops, a trend that has been observed since 2014. In contrast, newer datasets (i.e., \citet{DM-ASTE,Dom-Exp-ASTE}) address domain diversity by including home appliances, fashion, groceries, and more, moving beyond the typical restaurant and laptop reviews. Yet, these innovative datasets lack the critical aspect category annotations required for a holistic joint detection of all elements.

        \begin{table*}[t] 
        \centering
        
        \begin{minipage}{0.53\textwidth}
            \centering
            \rowcolors{2}{gray!25}{white}
            \resizebox{\columnwidth}{!}{%
            \begin{tabular}{lcccc}
            \hlineB{2}
            \rowcolor{gray!35}
            \textbf{Dataset} & \textbf{Tgt. Span} & \textbf{Opi. Ph. Span} & \textbf{Asp-Sen} & \textbf{Quadruple} \\ \hline
            Amazon\_FF & 72.57 & 69.72 & 85.43 & 65.78 \\
            Coursera   & 78.26 & 71.26 & 79.63 & 68.56 \\
            Hotels     & 74.78 & 72.05 & 87.32 & 73.84 \\ \hline
            \end{tabular}%
            }
        \end{minipage}\hfill 
        \begin{minipage}{0.40\textwidth}
            \centering
            \rowcolors{2}{gray!25}{white}
            \resizebox{\linewidth}{!}{%
            \begin{tabular}{lrrrrr}
            \hlineB{2}
            \rowcolor{gray!35}
            \multicolumn{1}{c}{\textbf{Dataset}} &
              \multicolumn{1}{c}{\textbf{\#sents}} &
              \multicolumn{1}{c}{\textbf{\#pos}} &
              \multicolumn{1}{c}{\textbf{\#neg}} &
              \multicolumn{1}{c}{\textbf{\#neu}} &
              \multicolumn{1}{c}{\textbf{\#Total}} \\ \hlineB{2}
            Rest-15 & 1,562 & 1,710          & 701           & 85           & 2,496          \\ 
            Rest-16 & 2,024 & 2,293          & 877           & 125          & 3,295          \\ \hline
            \textbf{Total} & \textbf{3,586} & \textbf{4,003} & \textbf{1,578} & \textbf{210} & \textbf{5,791} \\ \hlineB{2}
            \end{tabular}%
            }

        \end{minipage}\hfill
        \caption{\textbf{(a)} \textbf{Left:} Inter-Annotator agreement F1-scores for the OATS Datasets. Tgt. Span: Target span extraction. Opi. Ph. Span: Opinion Phrase span extraction. Asp-Sent: Aspect Category and Sentiment combination categorization. \textbf{(b)} \textbf{Right:} Current ASQP Dataset Statistics}
        \label{tab: IAA-Curr-Stats}
    \end{table*}

    Several crucial insights emerge in delving deep into the landscape of ABSA datasets \cite{sampreeth-ABSA-Datasets-Survey-arxiv}. Firstly, there needs to be more consistency in how ABSA components are defined and structured across various sources, accentuating an urgent need for a universal standardized format. Secondly, the challenge of accurately detecting opinion words, which are central to discerning sentiment polarity and their corresponding aspect terms, still needs to be solved. For instance, in the hotel review \textit{'the room was spotless and large, bathroom and kitchen were fine for our needs and the king size bed was great'}, the word 'spotless' indicates a positive sentiment polarity for the aspect 'rooms cleanliness', while 'large' pertains to 'rooms design\_features'. Correctly associating these opinion words with their respective aspects and targets is needed. Thirdly, the current datasets, while extensive, do not always reflect real-world complexities. 
    Different domains, such as healthcare, education, or entertainment, have their unique terminologies, sentiment expressions, and contextual nuances. 
    Moreover, the prevailing focus on English in ABSA datasets sidelines many low-resource languages, neglecting challenges tied to idiomatic expressions, linguistic structures, and cultural contexts inherent to these languages. 
    This need is further emphasized by recent advancements in unified models using generative frameworks \cite{T5-ABSA-sampreeth,T5-ABSA-towards-generative}, highlighting the immense potential of models that jointly solve ABSA tasks. These models necessitate datasets with richer annotations, similar to \cite{ASQP}. 
    Lastly, current ABSA research largely adopts a sentence-centric approach, risking misinterpretation by neglecting the inter-sentence context in reviews. While one might suggest merging sentence-level predictions, this doesn't remedy the foundational issue: sentiments in one sentence can be influenced by others. Without the full review context, even merged predictions can be off-mark. Although \cite{DM-ASTE} contains complete reviews of up to 250 words, they address only the ASTE task. Thus, there's a clear need for additional review-level ABSA datasets.
    
    Given these challenges, the OATS dataset has been developed with a vision to rectify most of the existing gaps. The contributions of OATS are:
    \begin{itemize}
        \item     Introduces both sentence-level quadruples and review-level tuples, capturing sentiments in all their granularity.
        \item     The dataset spans multiple domains, ensuring its applicability across a diverse range of ABSA tasks.
        \item     OATS will be publicly available in two formats: XML (as introduced by \citet{14-dataset} catering to detailed character-level annotations), and Text (following the format set by \citet{ASQP}).
        \item We provide baselines for our dataset, focusing on four primary tasks (ASD, ASTE, TASD, and ASQP) from the 14 ABSA subtasks.

    \end{itemize}

%% file: Sections/2-Related-Works.tex
In this section, we describe the tasks of ABSA and their related datasets, followed by triplet and quadruple extraction task-specific methods.

\subsection{Current Datasets}
Traditionally, when ABSA was first introduced by \citet{Hu_2004} as a task in NLP, the main aim then was to extract different aspects given a sentence followed by assigning polarities given those aspects. 
As a decade passed, \citet{14-dataset} added another important element of ABSA: aspect categories. In the SemEval-2014 shared task on ABSA, two sub-tasks were drafted in addition to the aspect term extraction and its sentiment polarity. They are the aspect category detection and their sentiment polarity. Two datasets from the restaurant and laptop domains were released as part of this shared task. 

In the consecutive years, \citet{15-dataset,16-dataset} proposed two more ABSA shared tasks at the SemEval, redefining the subtasks of ABSA and giving them a more concrete structure. 
Aspect categories were a combination of entity and attribute pairs, aspect terms were called targets, and the sentiment polarity was assigned jointly for the targets and the aspect categories. 
Several datasets with the three elements from multiple languages and domains were published, including English, Dutch, Spanish, French, Russian, and Turkish restaurants, English laptops, Arabic hotels, and many others. 

Later, \citet{TargetAspectSentimentJD} utilized the data from the SemEval-2015 and SemEval-2016 to define the target-aspect-sentiment joint detection task (TASD). After that, researchers felt the importance of detecting opinion phrases to identify the sentiment polarity and understand its relationship with targets and aspect categories. 
This resulted in three new sub-tasks with new datasets that stemmed from the SemEval challenges, namely aspect-opinion-pair extraction (AOPE) \cite{AOPE}, target-opinion-word extraction (TOWE) \cite{TOWE}, and aspect-sentiment-triplet-extraction (ASTE) \cite{ASTE-dataset}. 
More recently, the ABSA research has shifted towards the joint detection of all four elements of ABSA, which proved to better identify the inter-relationships among the elements, thereby enhancing the performance of other subtasks. 
This task is called aspect sentiment quadruple prediction (ASQP) \cite{ASQP} or aspect-category-opinion-sentiment joint detection (ACOS) \cite{ACOS}. The ASQP task is evaluated on the dataset with quadruples created from the SemEval challenges, ASTE triplets, and TOWE and AOPE tuples. 
The ACOS task introduced two datasets from the restaurant and the laptop domain with implicit and explicit aspect terms and opinion phrases. 

The current ABSA dataset landscape is notably skewed towards particular domains such as restaurants and electronics. This stems from the easy availability and volume of data on review websites and other online platforms. 
This narrow focus restricts the wider application of ABSA, limiting its ability to deliver insights across different sectors. 
Expanding the diversity of ABSA datasets is essential to push research and applications beyond the boundaries of mere reviews. To address this limitation, we propose three new datasets for the quadruple extraction task from three new domains, along with implicit and explicit targets and opinion phrase annotations. We also include the review-level tuples for identifying the overall sentiment polarity for different aspect categories in a review. This dataset could be used to solve all the above-mentioned subtasks of ABSA. 

\subsection{Related Methods}
We focus on a few joint detection tasks: the TASD, ASTE, and ASQP for our research. \citet{sampreeth-ABSA-Datasets-Survey-arxiv} have provided a detailed survey of the other datasets, tasks, and their challenges. 

\textbf{Triplet Extraction Tasks} 
In ASTE research, three primary methodologies have emerged: MRC-based techniques \cite{BMRC}, methods anchored on BERT and table-filling \cite{BDTF}, and generative approaches \cite{gen-scl-nat,GAS-T5,ASQP}. The MRC-based methods involve crafting a specific query for each component in the triplet, subsequently extracting them based on the response to this query. Generative methods, in contrast, frame the ASTE challenge as a sequence generation task and employ sequence-to-sequence (seq2seq) models. The triplets are then decoded using a specially tailored algorithm. In this study, we employ representative techniques from these three categories and investigate OATS using these methods. 

For the TASD task, there are two paths: BERT-based methods \cite{TargetAspectSentimentJD}, and generative-based approaches \cite{T5-ABSA-sampreeth,ASQP}. The BERT-based methods extract the aspect sentiment tuple from a sentence using the sentence-pair classification as a backbone \cite{14-Utilizing-Bert} followed by extracting the targets for each pair using the token classification with the BIO or a softmax classifier. The generative-based approaches convert the task similar to an abstractive summarization and use the sequence-to-sequence models like the T5, BART, and others \cite{T5,bart} to predict the triplets. 

\textbf{Quadruple Extraction Task}
Researchers have pointed out two promising directions. \citet{ACOS} propose a two-stage method by extracting the aspect and opinion terms first. Then, these items are utilized to classify aspect categories and sentiment polarity. Another method is based on the generation model \cite{ASQP}. By paraphrasing the input sentence, the quadruplet can be extracted end-to-end. In this work, we follow the generative direction and consider the order-free property of the quadruplet. A few other studies also proposed generative-based models with this paraphrasing as a backbone, including \citet{Template-Quad,gen-scl-nat}. 

%% file: Sections/3-Dataset.tex
In this section, we start by discussing the sources for the OATS dataset. The following subsections discuss the annotation process, including inter-annotator agreement, relevant data statistics, and OATS relevance to ABSA and the wider NLP community.


    \begin{table*}[t] 
        \centering
        \begin{minipage}{0.5\textwidth}
            \centering
        \rowcolors{2}{gray!25}{white}
        \resizebox{\linewidth}{!}{%
        \begin{tabular}{lrrrrr}
        \hlineB{2}
        \rowcolor{gray!35}
        \multicolumn{1}{c}{\textbf{Domain}} &
          \multicolumn{1}{c}{\textbf{\#Revs.}} &
          \multicolumn{1}{c}{\textbf{\#Sent.}} &
          \multicolumn{1}{c}{\textbf{\#Rev.Op.}} &
          \multicolumn{1}{c}{\textbf{\#Sent.Op.}} &
          \multicolumn{1}{c}{\textbf{\#Total Op}} \\ \hlineB{2}
            Amazon\_FF     & 1,794          & 8,913           & 4,326           & 8,260           & 12,586          \\
            Coursera       & 1,702          & 8,278           & 5,350           & 7,875           & 13,225           \\
            Hotels         & 1,497          & 7,963           & 7,416           & 11,335           & 18,751          \\ \hline
            \textbf{Total} & \textbf{4,993} & \textbf{25,154} & \textbf{17,092} & \textbf{27,470} & \textbf{44,562} \\ \hlineB{2}
            \end{tabular}%
            }
        \end{minipage}\hfill 
        \begin{minipage}{0.5\textwidth}
            \centering
            \rowcolors{2}{gray!25}{white}
            \resizebox{0.9\linewidth}{!}{%
            \begin{tabular}{crrr}
            \hlineB{2}
            \rowcolor{gray!35}
            \textbf{Stats/Domain} & \multicolumn{1}{c}{\textbf{Amazon\_FF}} & \multicolumn{1}{c}{\textbf{Coursera}} & \multicolumn{1}{c}{\textbf{Hotels}} \\ \hlineB{2}
            {Avg. Sentences/Review}   & 4.96        & 4.81         & 5.31         \\ 
            {Avg. Length of Sentence} & 71.34       & 79.02       & 75.95       \\ 
            {Avg. Length of Review}   & 359.9       & 391.55      & 405.29      \\ 
            {Avg. Opinions/Sentence}  & 0.92 (1.25) & 0.95 (1.27) & 1.42 (1.71)  \\ 
            {Avg. Opinions/Review}    & 2.4 (2.48)  & 3.16 (3.18) & 4.96 (5.37) \\ \hlineB{2}
            \end{tabular}%
            }
        \end{minipage}
        \caption{\textbf{(a) Left:} OATS Dataset Overall Statistics \textbf{(b) Right:} OATS Average Statistics. The values inside the () are the average opinions per sentence/review, excluding those with zero opinions. }
        \label{tab: OATS-Overall-Avg}
    \end{table*}

        We derived three distinct English review datasets from multiple sources.
        Our corpus is primarily built from text reviews available on Kaggle (consistent with the CCO and ODbL licenses) and resources provided by \citet{BeerAdvocate-TripAdvisor-dataset}.
        
        \paragraph{Amazon Fine Foods Dataset} This dataset, referred as Amazon\_FF in this work, is extracted from a Kaggle competition 
        \footnote{\url{https://www.kaggle.com/snap/amazon-fine-food-reviews}} 
        containing around 500k reviews of fine foods from Amazon. These reviews span topics such as the quality of food or products, promptness of delivery, packaging standards, and product availability, among others. 
        We curated 1,794 complete reviews from this dataset, consisting of over 8,900 sentences and approximately 8,200 opinion quadruples.
        
        \paragraph{Coursera Dataset} Originating from a Kaggle competition, this dataset contains reviews scraped from the Coursera website, totaling close to 100k reviews. These reviews reflect diverse perspectives on the course's quality, content, comprehensiveness, and the alignment of faculty lessons with course content. 
        For our work, we selected 1,702 comprehensive reviews, which include roughly 8,200 sentences and about 7,800 opinion quadruples.
        
        \paragraph{TripAdvisor Dataset} This dataset is based on data from \citet{BeerAdvocate-TripAdvisor-dataset}, featuring over 100k hotel reviews. The reviews comment on various facets of the hotel experience, such as pricing, design attributes, and more. 
        We collated 1,497 complete reviews from this extensive collection, resulting in approximately 8,000 sentences and nearly 11,300 opinion quadruples.

    \subsection{Annotation Procedure}
        We relied on Upwork to identify three freelancers for the data annotation and employed the BRAT tool for annotation. We provided a comprehensive guideline document introducing them to the topic, dataset, the BRAT tool, and specific annotation requirements. 
        Initially, annotator (A) took the lead in annotating a subset of the data (50 complete reviews), which was subsequently reviewed by another annotator (B) for corrections. 
        This process was iterated twice, with pairs of annotators (A, C) and then (B, C) for checks and balances. We used 150 reviews (50 for each annotator) from each domain for this phase of the agreement task. 
        Any emerging disagreements were addressed and resolved through consultations with one of our NLP experts, which are added as additional instructions to the annotators for future reviews. 
        
        This process is repeated multiple times until a decent inter-annotator agreement is reached among the three annotators. Then, the remaining dataset reviews were equally divided among the three annotators for the actual annotation process. 
        When all three annotators held differing views, a consensus was achieved in collaboration with an expert annotator. Gold annotations are generated via majority voting. 
        To ensure the high-quality of the annotations, for every 150 reviews, each annotator's work is reviewed by another annotator. If the agreement score is less than the initial score, that part of the data is re-annotated again until the score crosses the baseline. 

        According to \citet{not-kappa}, the infamous Kappa metric \cite{token-kappa1,token-kappa2} may not be the best fit for span-extraction annotation in textual data to measure the inter-annotator agreement. The limitation arises from the requirement of Kappa to compute the number of negative cases, which is unidentifiable for spans as they constitute sequences of words without a predetermined quantity of items for annotation in a text. 
        The F-measure is often more suitable for gauging inter-annotator agreement in span extraction annotation tasks such as target and opinion phrase extraction \cite{IAA-F1}. 
        We can calculate the F1-score by considering one annotator's annotations as the reference (or gold annotations or groundtruth) and the other's as a system's responses (or predictions). 
        
        We computed the inter-annotator agreement scores, following this F1 metric, for each domain using several combinations of the ABSA elements as shown in Table \ref{tab: IAA-Curr-Stats}(a). The average quadruple extraction agreement F1 score for the three domains is 69.39\%. 
        When we observe the simple task of aspect category and sentiment polarity identification, the average agreement F1-score is 84.12\%. If we go to challenging ones that include span extraction like the target and opinion phrase extraction, 
        This score is significant in light of the inherent subjectivity and complexity involved in jointly identifying all four ABSA elements. This highlights the considerable level of consistency among our annotators and, on the other hand, indicates the task's complexity.



    \begin{table*}[t] 
        \centering
        \begin{minipage}{0.58\textwidth}
            \centering
            \resizebox{0.95\columnwidth}{!}{%
            \rowcolors{2}{gray!25}{white}
            \begin{tabular}{l|rrrrr|rrrrr}
            \hlineB{2}
            \rowcolor{gray!35}
            \multicolumn{1}{c|}{} &
              \multicolumn{5}{c|}{\textbf{Sentence-Level}} &
              \multicolumn{5}{c}{\textbf{Review-Level}} \\ \cline{2-11} 
            \rowcolor{gray!35}
            \multicolumn{1}{c|}{\multirow{-2}{*}{\textbf{Domain}}} &
              \multicolumn{1}{c|}{\textbf{0-Op}} &
              \multicolumn{1}{c|}{\textbf{1-Op}} &
              \multicolumn{1}{c|}{\textbf{2-Op}} &
              \multicolumn{1}{c|}{\textbf{3-Op}} &
              \multicolumn{1}{c|}{\textbf{\textgreater{}3-Op}} &
              \multicolumn{1}{c|}{\textbf{0-Op}} &
              \multicolumn{1}{c|}{\textbf{1-Op}} &
              \multicolumn{1}{c|}{\textbf{2-Op}} &
              \multicolumn{1}{c|}{\textbf{3-Op}} &
              \multicolumn{1}{c}{\textbf{\textgreater{}3-Op}} \\ \hlineB{2}
            Amazon\_FF &
              \multicolumn{1}{r}{1,957} &
              \multicolumn{1}{r}{4,529} &
              \multicolumn{1}{r}{890} &
              \multicolumn{1}{r}{170} &
              42 &
              \multicolumn{1}{r}{47} &
              \multicolumn{1}{r}{175} &
              \multicolumn{1}{r}{626} &
              \multicolumn{1}{r}{490} &
              183 \\
            Coursera &
              \multicolumn{1}{r}{1,395} &
              \multicolumn{1}{r}{3,685} &
              \multicolumn{1}{r}{690} &
              \multicolumn{1}{r}{134} &
              36 &
              \multicolumn{1}{r}{8} &
              \multicolumn{1}{r}{107} &
              \multicolumn{1}{r}{298} &
              \multicolumn{1}{r}{349} &
              449 \\ 
            Hotels &
              \multicolumn{1}{r}{1,454} &
              \multicolumn{1}{r}{2,910} &
              \multicolumn{1}{r}{1,167} &
              \multicolumn{1}{r}{499} &
              337 &
              \multicolumn{1}{r}{116} &
              \multicolumn{1}{r}{11} &
              \multicolumn{1}{r}{58} &
              \multicolumn{1}{r}{138} &
              883 \\ \hline
            \textbf{Total} &
              \multicolumn{1}{r}{\textbf{4,806}} &
              \multicolumn{1}{r}{\textbf{11,124}} &
              \multicolumn{1}{r}{\textbf{2,747}} &
              \multicolumn{1}{r}{\textbf{803}} &
              \textbf{415} &
              \multicolumn{1}{r}{\textbf{171}} &
              \multicolumn{1}{r}{\textbf{293}} &
              \multicolumn{1}{r}{\textbf{982}} &
              \multicolumn{1}{r}{\textbf{977}} &
              \textbf{1,515} \\ \hlineB{2}
            \end{tabular}%
            }
        \end{minipage}\hfill 
        \begin{minipage}{0.42\textwidth}
            \centering
            \rowcolors{2}{gray!25}{white}
            \resizebox{\columnwidth}{!}{%
            \begin{tabular}{l|rrr|rrrr}
            \hlineB{2}
            \rowcolor{gray!35}
            \multicolumn{1}{c|}{} &
              \multicolumn{3}{c|}{\textbf{Sentence-Level}} &
              \multicolumn{4}{c}{\textbf{Review-Level}} \\ \cline{2-8} 
            \rowcolor{gray!35}
            \multicolumn{1}{c|}{\multirow{-2}{*}{\textbf{Domain}}} &
              \multicolumn{1}{c|}{\textbf{\#pos}} &
              \multicolumn{1}{c|}{\textbf{\#neg}} &
              \multicolumn{1}{c|}{\textbf{\#neu}} &
              \multicolumn{1}{c|}{\textbf{\#pos}} &
              \multicolumn{1}{c|}{\textbf{\#neg}} &
              \multicolumn{1}{c|}{\textbf{\#neu}} &
              \multicolumn{1}{c}{\textbf{\#conf}} \\ \hlineB{2}
            Amazon\_FF &
              \multicolumn{1}{r}{5,577} &
              \multicolumn{1}{r}{1,187} &
              234 &
              \multicolumn{1}{r}{2,900} &
              \multicolumn{1}{r}{606} &
              \multicolumn{1}{r}{74} &
              82 \\ 
            Coursera &
              \multicolumn{1}{r}{4,403} &
              \multicolumn{1}{r}{1,008} &
              213 &
              \multicolumn{1}{r}{2,910} &
              \multicolumn{1}{r}{721} &
              \multicolumn{1}{r}{129} &
              71 \\ 
            Hotels &
              \multicolumn{1}{r}{6,952} &
              \multicolumn{1}{r}{1,207} &
              169 &
              \multicolumn{1}{r}{4,557} &
              \multicolumn{1}{r}{817} &
              \multicolumn{1}{r}{110} &
              53 \\ \hline
            \textbf{Total} &
              \multicolumn{1}{r}{\textbf{16,932}} &
              \multicolumn{1}{r}{\textbf{3,402}} &
              \textbf{616} &
              \multicolumn{1}{r}{\textbf{10,367}} &
              \multicolumn{1}{r}{\textbf{2,144}} &
              \multicolumn{1}{r}{\textbf{313}} &
              \textbf{206} \\ \hlineB{2}
            \end{tabular}%
            }
        \end{minipage}
        \caption{\textbf{(a) Left:} OATS Dataset Statistics for the total number of sentences/reviews with the respective number of opinions \textbf{(b) Right:} OATS Dataset Statistics for the total number of positive, negative, neutral, and conflict sentiment polarities observed for the aspect categories at Review-Level and the aspect terms, aspect category, and opinion phrase triplets at the Sentence-Level in each domain.}
        \label{tab: OATS-sentiment-stats}
    \end{table*}

\subsection{The OATS dataset in Numbers}
\label{SubSec: stats}
    In Table \ref{tab: OATS-Overall-Avg} (a), we present some basic statistics of each domain in the datasets. It includes information about the total number of reviews, sentences, review-level opinion tuples, and sentence-level opinion quadruples for every domain. 
    Table \ref{tab: OATS-Overall-Avg} (b) has the averaged statistics, such as the sentences per review, length of sentence (and review), and the number of opinions per sentence (and review) with the domains as columns. 
    The numbers provided inside ``()'' for the number of opinions per sentence and review are computed after excluding the zero opinion sentences and reviews. 
    We also provide fine-grained statistics in Table \ref{tab: OATS-sentiment-stats} (a), such as the number of sentences and reviews with zero, one, two, three, and more than three opinions. 
    The stats corresponding to the number of opinions with different sentiment polarities, including positive, negative, neutral, and conflict at both the sentence and review levels, are presented in Table \ref{tab: OATS-sentiment-stats} (b).

    The hotels domain has a higher number of opinions per sentence as well as per review when compared to the other two domains. 
    Despite having the most reviews (from Table \ref{tab: OATS-Overall-Avg} (a)), the fine foods domain has the least number of opinions per sentence and review among the three. 
    One of the main reasons is that the number of zero opinion sentences in the fine foods domain is higher than in the hotels and Coursera (from Table \ref{tab: OATS-sentiment-stats} (a)). 
    Furthermore, the ratio of two-opinion sentences to one-opinion sentences for the Amazon\_FF and the Coursera domains ($\approx 1:5$) pushed the average numbers closer to one opinion per sentence. 
    We initially filtered the reviews for this dataset with a requirement of covering a minimum of 2 distinct aspect categories at the review level 
    is clearly evident from the average number of opinions per review. However, it was lacking at the sentence level.

    \begin{figure}[!h]
            \centering
            \includegraphics[width=\linewidth]{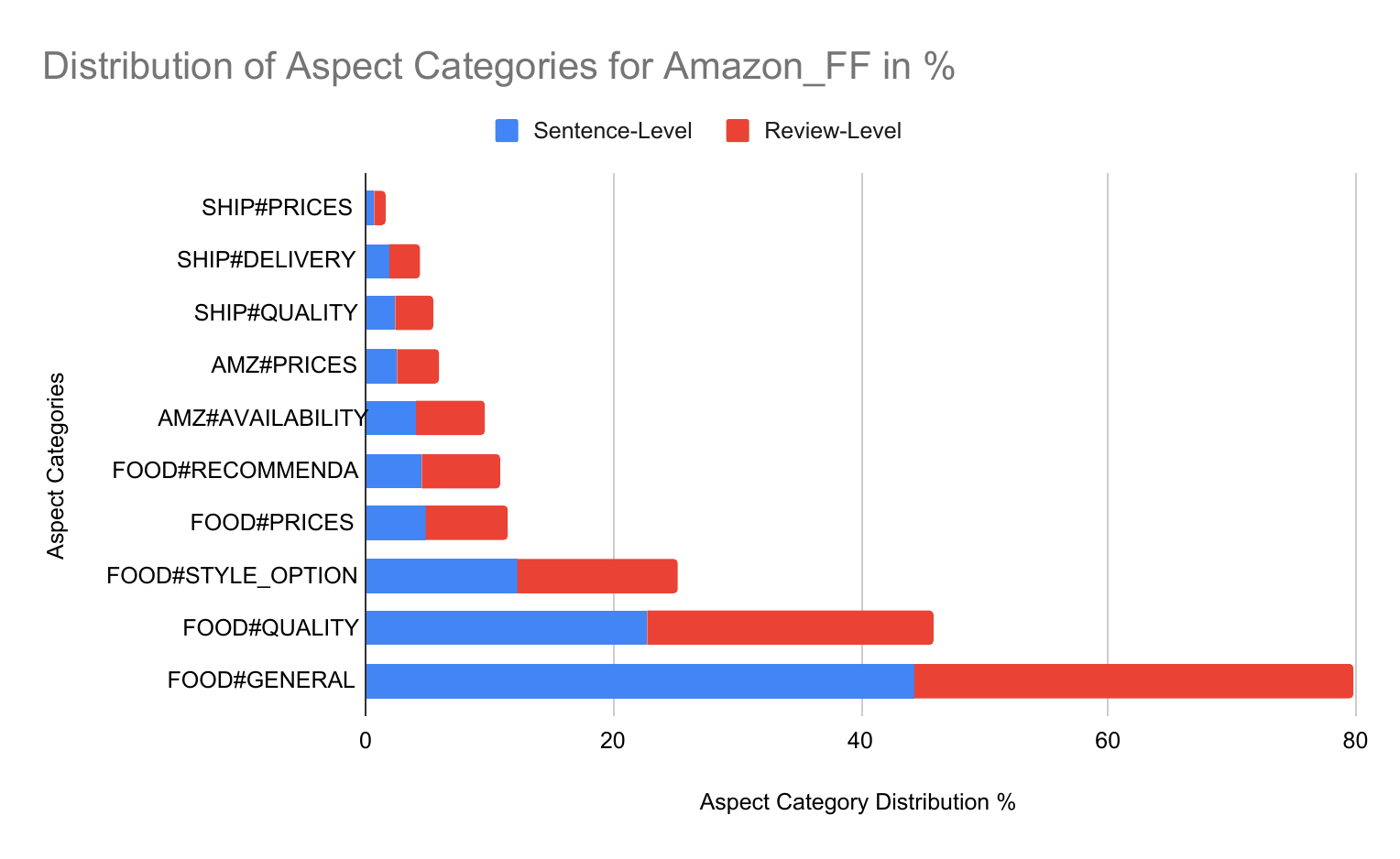}
            \caption{Distribution of Aspect Categories for Sentence-Level and Review-Level Annotations for Amazon\_FF Domain}
            \label{fig: OATS_asp_cat_distrb_amazon}
        \end{figure}
    
    Another important factor for ABSA datasets is the distribution of aspect categories. 
    Figure \ref{fig: OATS_asp_cat_distrb_amazon} shows that distribution for amazon\_FF domain. 
    The \textit{General} attribute for different entities in all the datasets has the highest number of opinions, such as the \textit{FOOD\#GENERAL} in the Amazon\_FF, \textit{COURSE\# GENERAL} and \textit{FACULTY\#GENERAL} for the Coursera domain, and \textit{HOTEL\#GENERAL}, \textit{LOCATION\#GENERAL}, and \textit{SERVICE\#GENERAL} in the hotel domain. 
    The review-level and the sentence-level opinions have similar distributions for the aspect categories. 
    We observed a similar distribution among the aspect categories in the other domains, so we chose to present only one domain plot for representation.



%% file: Sections/4-Experiments.tex
This section will first list the tasks we conduct experiments on, followed by the baseline methods for each task. We divide the methods into task-specific and unified approaches. Finally, we discuss the evaluation metrics for the tasks. 


    \begin{table*}[t] 
        \centering
        \begin{minipage}{0.48\textwidth}
            \centering
            \rowcolors{2}{gray!25}{white}
            \resizebox{0.9\columnwidth}{!}{%
            \begin{tabular}{lrrr}
            \hlineB{2}
            \rowcolor{gray!30}
            \textbf{Method} & \multicolumn{1}{l}{\textbf{Amazon\_FF}} & \multicolumn{1}{l}{\textbf{Coursera}} & \multicolumn{1}{l}{\textbf{Hotels}} \\ \hlineB{2}
            GAS-T5       & 19.62          & \textbf{22.23} & 26.33          \\
            Paraphrase   & 20.84          & 19.78          & \textbf{34.51} \\
            Template-ILO & \textbf{21.01} & 21.24          & 24.62          \\
            Template-DLO & 20.39          & 21.96          & 26.59          \\
            GEN-SCL-NAT  & 20.36          & 20.20          & 28.58          \\ \hlineB{2}
            \end{tabular}%
            }
        \end{minipage}\hfill 
        \begin{minipage}{0.48\textwidth}
            \centering
            \rowcolors{2}{gray!25}{white}
            \resizebox{0.9\columnwidth}{!}{%
            \begin{tabular}{lrrr}
            \hlineB{2}
            \rowcolor{gray!30}
            \textbf{Method} & \multicolumn{1}{l}{\textbf{Amazon\_FF}} & \multicolumn{1}{l}{\textbf{Coursera}} & \multicolumn{1}{l}{\textbf{Hotels}} \\ \hlineB{2}
            BDTF (BERT) & 27.81          & 43.98          & 42.02          \\
            BMRC                         & \textbf{49.48} & \textbf{50.85} & \textbf{50.71} \\ 
            GEN-SCL-NAT                  & 20.36          & 20.20          & 28.58          \\
            GAS                          & 46.46          & 40.15          & 45.96          \\
            Paraphrase                   & 46.64          & 39.30          & 46.37          \\
            \hlineB{2}
            \end{tabular}%
            }
        \end{minipage}
        \caption{\textbf{(a) Left:} F1 scores of ASQP task on OATS  \textbf{(b) Right:} F1 scores of ASTE task on OATS. Best results are highlighted in bold font.}
        \label{tab: ASQP-ASTE}
    \end{table*}

\subsection{Tasks}
\label{SubSec: Tasks}
We selected three major joint detection tasks for our sentence-level ABSA experiments: target-aspect-sentiment detection (TASD) \cite{TargetAspectSentimentJD}, aspect sentiment triplet extraction (ASTE) \cite{ASTE-dataset}, and aspect sentiment quadruple prediction (ASQP) \cite{ASQP}. The rationale for choosing these tasks is that they encompass all four elements of ABSA in different combinations. Our emphasis on joint extraction tasks over single-element extraction is motivated not just by the findings of \citet{T5-ABSA-sampreeth}, but also by broader research trends in the NLP community. For instance, the DREAM paper on entity-relation extraction successfully illustrates the advantages of jointly modeling evidence ranking, leading to enhanced performance \cite{DREAM-paper}. Such joint models, tailored to manage multiple intertwined elements simultaneously, have consistently been shown to surpass models fine-tuned for single-element extraction in various domains.

For the review-level (or text-level) ABSA, where we have to extract aspect category and sentiment tuples, we chose to use the aspect sentiment joint detection task (ASD) \cite{14-Utilizing-Bert} for our experiments. 
The two main challenges with the review-level task are that we have to use the entire review context rather than simply using each sentence to detect the aspect categories and the sentiment polarity of the review. Secondly, there can be multiple opinions on the same aspect category with multiple polarities, from which we must assign the dominating sentiment to each category. 
In the review, ``The room was mostly spotless and the bathroom pristine, but there was dust on the bedside table," two positive sentiments outweigh a milder negative one for the room's cleanliness, leading to an overall positive polarity.

\subsection{Baseline Methods}
    We implement several representative models from various frameworks, including MRC-based approaches similar to \citet{MRC-based1,MRC-based2,MRC-based3}, generation-based, and BERT-based frameworks \cite{BERT}, for the task evaluations.

    \paragraph{Task-specific methods} 

        For \textit{ASTE}, we use the following methods:

        \textbf{BMRC} \cite{BMRC}: a MRC-based method. It extracts aspect-oriented triplets and opinion-oriented triplets. Then, it obtains the final results by merging the two directions. 


        \textbf{BDTF (BERT)} \cite{BDTF}: a BERT-based method that uses table-filling to solve the problem. It transforms the ASTE task into detecting and classifying relation regions in the 2D table representing each triplet in addition to an effective relation representation learning approach to understand word and relation interactions. 

        These are the methods for \textbf{ASQP} task:
        \textbf{Template-ILO} \cite{Template-Quad}: a generative-based method.  Similar to \citet{ASQP} with an additional step to identify the best permutation of the four ABSA elements at the \textit{instance level}, and further combine multiple proper templates as data augmentation, instead of having a fixed order template, that is passed as input to the generative model. 

        \textbf{Template-DLO} \cite{Template-Quad}: a generative-based method. Similar to \citet{ASQP}, but has an additional step to identify the best permutation of the four ABSA elements at the \textit{dataset level}, and further combine multiple proper templates as data augmentation, instead of having a fixed order template, that is passed as input to the generative model.

        \paragraph{TASD Task-specific methods} 
        For the TASD task, we chose \textbf{TAS-BERT} \cite{TargetAspectSentimentJD}, which is a BERT-based method. It fine-tunes the pre-trained BERT model to solve the aspect-sentiment detection task using the classification token and then detects the targets corresponding to those tuples using the token classification with CRF/softmax decoding.

    \begin{table*}[t] 
        \centering
        \begin{minipage}{0.4\textwidth}
            \centering
            \rowcolors{2}{gray!25}{white}
            \resizebox{\columnwidth}{!}{%
            \begin{tabular}{llll}
            \hlineB{2}
            \rowcolor{gray!30}
            \textbf{Method} & \textbf{Amazon\_FF} & \textbf{Coursera} & \textbf{Hotels} \\ \hlineB{2}
            TAS-BERT-BIO        & \multicolumn{1}{r}{45.12}  &  \multicolumn{1}{r}{44.41}     &  \multicolumn{1}{r}{45.92}               \\
            TAS-BERT-TO        & \multicolumn{1}{r}{47.51}  &  \multicolumn{1}{r}{42.77}     &  \multicolumn{1}{r}{45.76}               \\
            T5-ABSA & \multicolumn{1}{r}{\textbf{51.61}} & \multicolumn{1}{r}{\textbf{44.57}}          & \multicolumn{1}{r}{49.78}          \\
            GAS        & \multicolumn{1}{r}{43.04}          & \multicolumn{1}{r}{41.53} & \multicolumn{1}{r}{\textbf{50.69}} \\
            Paraphrase & \multicolumn{1}{r}{44.89} & \multicolumn{1}{r}{40.24}          & \multicolumn{1}{r}{49.81}          \\ \hline
            \end{tabular}%
            }
        \end{minipage}\hfill 
        \begin{minipage}{0.58\textwidth}
            \centering
            \rowcolors{2}{gray!25}{white}
            \resizebox{0.9\columnwidth}{!}{%
            \begin{tabular}{lrrrrrr}
            \hlineB{2}
            \rowcolor{gray!35}
            \multicolumn{1}{c}{} &
              \multicolumn{2}{c}{\textbf{Amazon\_FF}} &
              \multicolumn{2}{c}{\textbf{Coursera}} &
              \multicolumn{2}{c}{\textbf{Hotels}} \\ \cline{2-7} 
            \rowcolor{gray!35}
            \multicolumn{1}{c}{\multirow{-2}{*}{\textbf{Method}}} & \textbf{R-ASD} & \textbf{S-ASD} & \textbf{R-ASD} & \textbf{S-ASD} & \textbf{R-ASD} & \textbf{S-ASD} \\ \hline
            BERT-pair-NLI-B & 88.22          & 93.25          & 91.31          & 95.92          & 91.84          & 96.82          \\
            BERT-pair-QA-B  & \textbf{88.42} & \textbf{93.61} & \textbf{91.51} & \textbf{96.69} & \textbf{91.93} & \textbf{97.18} \\
            QACG-BERT-NLI-M      & 87.15          & 92.13          & 90.41          & 95.81          & 90.05          & 95.68          \\
            T5-ABSA              & 58.26          & 56.85          & 40.76          & 48.81          & 54.09          & 59.85          \\
            GAS-T5               & 65.23          & 57.57          & 46.61          & 56.79          & 58.61          & 66.9           \\
            Paraphrase           & 66.46          & 57.25          & 47.88          & 55.27          & 56.11          & 66.38          \\ \hline
            \end{tabular}%
            }
        \end{minipage}
        \caption{\textbf{(a) Left:} F1 scores of in-domain TASD on OATS. \textbf{(b) Right:} Results for Aspect-Sentiment Joint Detection task at Review-level (R-ASD) and Sentence-level (S-ASD) on OATS datasets. Best results are highlighted in bold font.}
        \label{tab: TASD-R-S-ASD}
    \end{table*}

    \paragraph{ASD Task-specific methods} The below methods are used for both review-level (R-ASD) and sentence-level (S-ASD) tasks:
    



        \textbf{BERT-pair-NLI-B }\cite{14-Utilizing-Bert}: a BERT-based model that takes the review context as segment A and predicts the presence of the combination of aspect category and sentiment polarity, which is taken as an entailment in the segment B of BERT's input, using the output from [CLS] token. 
        
        \textbf{BERT-pair-QA-B} \cite{14-Utilizing-Bert}: a BERT-based model that uses question-answering format instead of entailment in BERT-pair-NLI-B. 

        \textbf{QACG-BERT} \cite{QACG-BERT}: developed a CG-BERT, which utilizes context-guided (CG) softmax-attention by initially modifying a context-aware Transformer. Subsequently, they introduced an improved Quasi-Attention CG-BERT model that learns a compound attention mechanism, facilitating subtractive attention.

        \textbf{T5-ABSA} \cite{T5-ABSA-sampreeth}: a generative approach that takes the review context along with a task prefix as input to generate the respective task outputs in an auxiliary sentence-based or phrase-based format defined in their work.

    \paragraph{Unified methods} 
        The following generative frameworks can be applied to any sentence-level and review-level ABSA tasks:

        \textbf{GEN-NAT-SCL} \cite{gen-scl-nat}: a generative-based method. It combines a new generative format with a supervised contrastive learning objective to predict the ASTE's triplets and ASQP's quadruples. 
        
        \textbf{GAS} \cite{GAS-T5}: a generation-based method. It transforms the ASTE, ASQP, TASD, and ASD tasks into a text generation problem that inputs the review and generates all the respective combinations of ABSA opinion elements as output. 
        
        \textbf{Paraphrase} \cite{ASQP}: a generative-based method. It is similar to GAS but transforms output opinion elements into paraphrases that read as natural sentences. We substituted the corresponding element for implicit targets and opinions with the word "it." 
    
\subsection{Evaluation}

    Following \citet{ASQP,DM-ASTE,TargetAspectSentimentJD}, we use the F1 score to measure the performance of different approaches on all the tasks from Section \ref{SubSec: Tasks}. All experimental results are reported using the average of 5 different runs using distinct random seeds. We divided each domain dataset into train, validation, and test sets with 80\%, 10\%, and 10\% splits, respectively. A tuple, triplet, and quadruple is considered correct only if all the corresponding prediction elements match the gold standard labels. We consider any partial matches as wrong predictions following \citet{ASQP}. We adopt the base versions of all the transformer models for our experiments, including the BERT-base \cite{BERT}, T5-base \cite{T5}, RoBERTa-base \cite{roberta}, and others.

%% file: Sections/5-Discussion.tex

        In this section, we present the results of different baseline systems on the selected tasks, highlighting the benefits and challenges of current methods on the OATS dataset and the specific task. 

         \textbf{ASQP} 
        Table \ref{tab: ASQP-ASTE} (a) reports the performance on the ASQP task for the quadruple extraction using five different baseline systems on the three datasets. On a high level, the overall performance of the hotels domain is better compared to the other two domains in all the baseline systems. 

        For the Amazon\_FF domain, even though Template-ILO gave the best performance, all the other methods are not far from it. Similarly, for the Coursera domain, GAS-T5 performed better than the Paraphrase and GEN-SCL-NAT, but the Template method is close to its performance. Paraphrase outperformed all the other methods in the hotels domain with a 34.51\% F1 score. 

        Prior works have shown that the Template method significantly outperformed Paraphrase and GAS in the restaurant domain \cite{Template-Quad}. That trend only translated to the Amazon\_FF domain, leaving out Coursera and Hotels. Similarly, GEN-SCL-NAT, which outperformed the Paraphrase method on the restaurants and laptops domain, failed to carry it to any of the three domains.


    \begin{table*}[t] 
        \centering
        \begin{minipage}{0.45\textwidth}
            \centering
            \rowcolors{2}{gray!25}{white}
            \resizebox{\columnwidth}{!}{%
            \begin{tabular}{lrrrrrr}
            \hlineB{2}
            \rowcolor{gray!35}
            \textbf{Domain/Task} &
              \multicolumn{1}{l}{\textbf{TOWE}} &
              \multicolumn{1}{l}{\textbf{TSD}} &
              \multicolumn{1}{l}{\textbf{ASD}} &
              \multicolumn{1}{l}{\textbf{TASD}} &
              \multicolumn{1}{l}{\textbf{ASTE}} &
              \multicolumn{1}{l}{\textbf{ASQP}} \\ \hline
            Amazon\_FF & 31.29 & 65.27 & 57.25 & 44.89 & 46.64 & 20.84 \\
            Coursera   & 38.64 & 67.75 & 55.27 & 40.24 & 39.30 & 19.78 \\
            Hotels     & 35.75 & 62.34 & 66.38 & 49.81 & 46.37 & 34.51 \\ \hline
            \end{tabular}%
            }
        \end{minipage}\hfill 
        \begin{minipage}{0.51\textwidth}
            \centering
            \rowcolors{2}{gray!25}{white}
            \resizebox{\columnwidth}{!}{%
                \begin{tabular}{lrrrrrr}
                \hlineB{2}
                \rowcolor{gray!35}
                \textbf{Domain} & \multicolumn{1}{l}{\textbf{ET / IT}} & \multicolumn{1}{l}{\textbf{EO / IO}} & \multicolumn{1}{l}{\textbf{ET-EO}} & \multicolumn{1}{l}{\textbf{ET-IO}} & \multicolumn{1}{l}{\textbf{IT-EO}} & \multicolumn{1}{l}{\textbf{IT-IO}} \\ \hline
                {Amazon\_FF} & 2,999 / 5,261 & 6,780 / 1,480 & 2,491 & 508 & 4,298 & 972 \\
                {Coursera} & 5,163 / 2,712 & 6,185 / 1,690 & 4,222 & 941 & 1,963 & 749 \\
                {Hotels} & 8,654 / 2,820 & 10,193 / 1,281 & 7,927 & 727 & 2,266 & 554 \\ \hline
                
                \end{tabular}
            }
        \end{minipage}
        \caption{\textbf{(a) Left:} Performance Analysis on six main ABSA joint tasks using the Paraphrase method \cite{ASQP} on OATS.  \textbf{(b) Right:} Distribution of Explicit and Implicit Targets (ET and IT) and Explicit and Implicit Opinion Phrases (EO and IO) in the OATS datasets with their different combinations.}
        \label{tab: Paraph-Imp-Exp-Op-Tgt}
    \end{table*}

        \textbf{ASTE}
        In Table \ref{tab: ASQP-ASTE} (b), we present the performance of five baseline systems on the ASTE task, which focuses on extracting the triplet of aspect term, opinion term, and sentiment polarity across three datasets. Notably, the BMRC method emerges as the standout performer, surpassing other baselines across all domains for this task, while the GEN-SCL-NAT method consistently lags behind. The success of the BMRC method can be attributed to the approach taken by \citet{BMRC}. Instead of solely considering the relationship from target to opinion phrases, they also incorporate the potential backward relation from opinion phrases to target expressions. This deviates from other methods like \citet{seq2path} and \citet{Opinion-Tree-Generation}, which deploy two distinct paths in a tree structure for targets and opinion phrases to identify the quadruples. 

    \textbf{R-ASD}
    Table \ref{tab: TASD-R-S-ASD} (b) shows the aspect sentiment joint detection task results at the review-level (R-ASD) and sentence-level (S-ASD). 
    BERT-based methods significantly outperform the generative approaches. One primary factor driving this difference is the augmentation strategy BERT-pair methods use, which expands the dataset based on the number of categories multiplied by the polarities in that dataset. For example, with 27 aspect categories and 4 different polarities in the Coursera domain for the ASQP task, this results in a total of 894024 instances ($8278 * 27 * 4$). Additionally, these methods transform the multi-label, multi-class classification of the ASD task into a simpler binary classification, making the task inherently easier than the challenge generative models face when extracting categories and polarities directly from a review sentence.

    \textbf{TASD} 
    We present the results of the TASD task experiments on all the domains in Table \ref{tab: TASD-R-S-ASD} (a). The generative-based models performed better than the BERT-based approaches on all the domain datasets. 
    The T5-ABSA method performed best for Amazon\_FF and Coursera domains, while the GAS method outperformed the hotels domain. All the generative-based approaches are not far from each other in the hotels domain, while the BERT-based methods have a significant difference in the Amazon\_FF and hotels domain when compared to the generative approaches.

    \textbf{Explicit and Implicit Targets and Opinions}
    We observe distinct patterns of explicit and implicit targets and opinions across the OATS domains from the statistics presented in Table \ref{tab: Paraph-Imp-Exp-Op-Tgt} (b). Amazon\_FF, for instance, showcases a predominance of implicit targets, suggesting users often leave their main subject of discussion implied. 
    Conversely, Coursera leans more towards explicit target mentions, possibly indicative of the direct nature of feedback in educational settings. 
    Notably, regardless of the domain, explicit opinions outnumber their implicit counterparts. This trend underlines that while users might leave their subjects (targets) implied, they tend to be direct about their sentiments or opinions on them.
    The Hotels domain stands out with the highest counts of explicit mentions for both targets and opinions, hinting at the straightforward nature of feedback in this domain. It is important to recognize these domain-specific trends as they aid in tailoring models appropriately, potentially improving their accuracy and adaptability. 

    \textbf{Challenging combination of elements in OATS}
    We tested the Paraphrase-T5 method \cite{ASQP} on all the OATS datasets for six different ABSA tasks to see how hard it is to detect various ABSA elements. The results are in Table \ref{tab: Paraph-Imp-Exp-Op-Tgt} (a). The TSD task did the best on all the domains, but TOWE was the hardest. This shows that connecting an opinion phrase to its target is especially tough in OATS. Interestingly, even though TOWE is kind of a part of ASTE, it does not do as well. This might be because having sentiment information in ASTE helps pick out opinion phrases better, giving it an edge. 
    
    Furthermore, integrating aspect categories into TSD creating the TASD task, saw a noticeable performance drop. This highlights the challenge of pinpointing aspect categories for specific targets. This challenge could also explain the lower scores of the ASD and ASQP tasks in the Amazon\_FF and Coursera domains compared to TSD. 
    Interestingly, in the hotels domain, ASD emerged as the top-performing task, which contributed to a superior TASD and ASQP performance relative to other domains. This success in the hotels domain might also explain the lower scores of the ASD and ASQP tasks in the Amazon\_FF and Coursera domains compared to TSD. 
    The complex nature of ASQP, which demands establishing relationships between categories, opinion phrases, and targets, might account for its sub-optimal performance stemming from the above two reasons.

%% file: Sections/5-Significance.tex
    The main differences between the current ASQP datasets and OATS are: \textbf{(a)} we annotated the data from scratch with all the elements together instead of aligning from several sources, \textbf{(b)} we include all the implicit aspect terms and opinion phrases, unlike ASQP, which excluded implicit opinion terms, \textbf{(c)} we provide both the review-level and sentence-level annotations facilitating the analysis and experiments for the review/text-level aspect-based sentiment analysis that differentiates \citet{MEMD-ABSA} from our dataset, \textbf{(d)} there are $\approx 5.8K$ opinion quadruples in total for the ASQP corpus, which includes two restaurant datasets that are just 80\% of opinion quadruples of the Coursera domain itself, which is the smallest in the OATS corpus. \textbf{(e)} OATS has a total of $\approx 27.5K$ opinion quadruples, almost five times larger than the ASQP corpus. Also, if we include the review-level annotations, we have nearly $44.5K$ opinions in OATS. 

    The OATS corpus will help analyze and understand the inherent relationships among all the elements of ABSA and exploit them to solve the ABSA task holistically. 
    The comprehensive nature of the OATS corpus not only allows the evaluation of the ASQP task but also unlocks the potential analysis and exploration of all the subtasks of ABSA, including TASD, ASTE, TOWE, TAD, and many more. 
    As pointed out by \cite{ASQP}, the characteristic of tackling various ABSA tasks in a unified framework enables the knowledge to be easily transferred across related subtasks, which is especially beneficial under low-resource settings. 
    It also allows cross-task transfer for subtasks that underperform when trained using only a task-specific dataset. 
    The provided review-level annotations will help many researchers who like to explore ABSA for whole reviews rather than individual sentences or single-sentence reviews. 

%% file: Sections/6-Conclusion.tex
ABSA still presents a lot of difficulties in the rapidly developing field of generative artificial intelligence. We present a thoroughly curated ABSA dataset including quadruples, both explicit and implicit attributes, and views, spanning three unique domains. In addition, the dataset incorporates review-level sentiment polarity for each aspect category, providing a comprehensive perspective of the sentiments expressed in the reviews. Our annotations outnumber those found in previous datasets. We do an in-depth study, give comprehensive annotation guidelines, and provide dataset statistics. We also enable evaluations of generative and non-generative benchmarks on a range of common ABSA tasks.

%% file: Sections/8-Ethics.tex
We identified the following limitations of this work:
\begin{enumerate}
    \item While the OATS dataset spans multiple domains, including the relatively unexplored domains of Amazon fine foods and Coursera course reviews, it exclusively contains English-language reviews. The dataset currently lacks reviews in low-resource languages. We plan to address this in future works by releasing multi-lingual datasets, inclusive of the low-resource setting.
    \item The experimental results showcased in this paper are purely in-domain; models were fine-tuned and tested within the same domain. In subsequent works, we aim to design experiments to tackle cross-domain and out-of-domain ABSA tasks.
    \item Our dataset incorporates review-level ABSA tuples focusing only on aspect categories and sentiment polarities. It currently omits review-level target expressions and opinion phrase annotations, which would render the dataset more comprehensive for ABSA. We earmark this as an avenue for future exploration.
\end{enumerate}

We point out the following ethical considerations while building and using the OATS dataset:
\begin{enumerate}
    \item The data collected from platforms such as Amazon Finefoods, Coursera, and TripAdvisor Hotels come from public reviews provided by users. We ensured that all identifiable information, including usernames, avatars, and any other potentially identifying details, were removed to preserve the anonymity of the reviewers.
    \item Our data processing methodology focused on extracting and analyzing the content of the reviews without altering the original sentiment or meaning. We handled this data with the utmost care to ensure that the sentiments and opinions of the original reviewers were not misrepresented.
    \item We recognize that reviews from online platforms might not represent the complete spectrum of users' opinions, as they may be influenced by various factors like platform algorithms, user demographics, and more. We urge users of this dataset to be aware of potential biases and always consider the data in the proper context.
    \item While the reviews were publicly available, the original authors might not have expected their content to be used in research. We've taken measures to respect their privacy, but future users of this dataset should also be aware of this consideration.
    \item While the reviews are publicly accessible, we acknowledge the platforms (Amazon Finefoods, Coursera, TripAdvisor Hotels) as the source of this data. We have only used this data for research and academic purposes, ensuring that we respect the terms of use for each platform.
\end{enumerate}